\documentclass[sigconf, nonacm]{acmart}

\newcommand{\hush}[1]{}

\usepackage{amsmath}
\DeclareMathOperator*{\argmax}{arg\,max}

\AtBeginDocument{%
  \providecommand\BibTeX{{%
    \normalfont B\kern-0.5em{\scshape i\kern-0.25em b}\kern-0.8em\TeX}}}

\setcopyright{acmcopyright}
\copyrightyear{2021}
\acmYear{2021}
\acmDOI{10.1145/1122445.1122456}

\DeclareMathOperator{\E}{\mathbb{E}}
\usepackage{amsmath,amsfonts}
\usepackage{algorithmic}
\usepackage{algorithm}
\usepackage{bm}
\usepackage{chngcntr}
\usepackage{graphicx}
\usepackage{textcomp}
\usepackage{xcolor}
\usepackage{cuted, nccmath}
\usepackage{lipsum}
\usepackage{mathtools}
\usepackage{subfigure}
\usepackage{graphicx}

\makeatletter
\setcopyright{none}
\makeatother

\begin{document}

\title{Personalization for Web-based Services \\
using Offline Reinforcement Learning}


\author{Pavlos Athanasios Apostolopoulos}
\email{pavlosapost@unm.edu}
\affiliation{%
  \institution{
The University of New Mexico, ECE
}
Albuquerque, NM
}

\author{Zehui Wang, Hanson Wang, Chad Zhou, 
Kittipat Virochsiri, Norm Zhou, Igor L. Markov}
\email{{wzehui,hansonw,yuzhoubrother,kittipat,nzhou,imarkov}@fb.com}
\affiliation{%
  \institution{Facebook Inc., Menlo Park, CA}
}

\hush{
\author{}
\email{@fb.com}
\affiliation{%
  \institution{Facebook Inc.}
  Menlo Park, CA
}
}

\renewcommand{\shortauthors}{Apostolopoulos et al.}

\begin{abstract}
   Large-scale Web-based services present opportunities for improving UI policies based on observed user interactions. We address challenges of learning such policies through model-free offline Reinforcement Learning (RL) with off-policy training. Deployed in a production system for user authentication in a major social network, it significantly improves long-term objectives. We articulate practical
   challenges, compare several ML techniques, provide insights on training and evaluation of RL models, and discuss generalizations.
\end{abstract}



\maketitle

\section{Introduction}
For Web-based services with numerous customers, such as social networks, UI decisions impact top-line metrics, such as user engagement, costs and revenues. Machine learning, especially Supervised Learning, can optimize these decisions but long-term cumulative objectives makes it challenging to label each decision for training. 
As an illustration, consider online user authentication. When a user mistypes or forgets her password, the service can loop back to the login prompt or offer another login channel, authorization code via SMS (Short Message Service), etc.
The user’s context is a part of a personalized configuration. The scale of the social network makes monetary cost (service fees for authentication) significant. Some users get their password right on the second try or look it up but others give up, and this is reflected in daily/monthly user engagement. Deciding when to authenticate requires real-time personalization with consideration for cost and engagement metrics, but the impact on end metrics is delayed. Our work addresses these challenges with ML, describes insights on model selection and training, and reports a production deployment. We also explain how our general approach and our infrastructure generalize to multiple challenges beyond the didactic application.

\noindent
{\bf Prior efforts} on personalized configuration systems covers a wide range of content-serving applications, including recommendation systems \cite{garcin2013personalized, garcin2013pen}, ad targeting \cite{tougucc2020hybrid, 10.1145/2556195.2556252}, and personalized medical healthcare \cite{ho2019machine}. Recent advances in Machine Learning brought us expressive deep neural networks with effective generalization, as well as infrastructure
for building real-time systems powered by ML. A common theme
is to {\em provide a personalized configuration and
optimize desired end metrics given user's real-time context}. Supervised Learning methods have shown promising results \cite{wang2018path, hidasi2015session}, but they only focus on optimizing immediate metrics such as the click-through rate \cite{li2017neural} and conversion rate \cite{pradel2011case}. This limitation is important for long-term user engagement ({\em daily/monthly active users} and monetary cost), as shown by the authentication example. 
\hush{Modeling long-term metrics within supervised learning can be conceptually difficult and also requires a greater engineering effort.}

\noindent
{\bf Reinforcement Learning (RL)} \cite{sutton2018reinforcement} seeks an optimal policy to maximize a long-term reward, and thus helps drive real-time personalized decisions \cite{tang2019reinforcement}. Here, the RL agent's {\em environment} is an individual online user, and the {\em personalized serving procedure} is modeled via sequential
agent-environment interactions. Therefore, the RL agent alternates between policy improvement and interaction with the environment. In practice, RL agents need many interactions to learn good policies \cite{dulac2019challenges}. To this end, training RL agents online may undermine user experience and/or incur large costs. Hence, policies are trained offline using logged interactions from any type of prior policies ({\em Offline RL} in
Section \ref{sec:offline}).

\noindent
{\bf In this work}, we use Offline RL to improve personalized authentication for a Web-based service. We formalize the problem as a Markov decision process (MDP) \cite{puterman2014markov}, where the RL agent learns a personalized policy, i.e., when to send an authentication message to the user after a failed login attempt. The training process optimizes long-term user engagement and the service's authentication costs. We avoid the pitfalls of online trial-and-error by training on prior experiences logged for different policies through {\em off-policy learning}. To ensure effective offline training,
we use several training heuristics, then evaluate performance of the learned candidate policies via an \textit{unbiased off-policy} estimator. The best learned policy is chosen and evaluated in online A/B tests \cite{kohavi2017online} on live data.
\hush{We share insights into modeling and offline training, and detail online evaluation to help practitioners train Reinforcement Learning agents on logged data.}

The remaining part of the paper is organized as follows. We review Reinforcement Learning in Section \ref{sec:background} focusing on off-policy learning. Section \ref{sec:offline} discusses the need for offline learning and the challenges of deploying such applications. A representative application amenable to reinforcement learning is covered in Section \ref{sec:auth}, and problem formalization in Section \ref{sec:formal}. 
Our contributions are introduced in Section \ref{sec:apply} where we describe an industry application of Offline RL to user authentication.
Section \ref{sec:empirical} details how our Offline RL model was deployed to a major social network and compares it to a supervised-learning baseline.
Section \ref{sec:conclusions} puts our work in perspective and discusses personalization more broadly.

\section{Background}
\label{sec:background}
In this section, we review the background and establish notation that will be used later.
Section \ref{subsec:rl} reviews Reinforcement Learning theory \cite{sutton2018reinforcement} and Section \ref{Off-policy} focuses on \textit{off-policy} learning.
\vspace{-0.3cm}
\subsection{Preliminaries on Reinforcement Learning} \label{subsec:rl}
Reinforcement learning seeks to control an interactive dynamic system. At each discrete time step, $t = 0, 1, \cdots$, the agent observes the environment's state $s_t$ and responds with action $a_t$, while the environment responds with an associated reward $r_{t+1}$ and transitions into the next state $s_{t+1}$. To this end, the environment can be defined by a Markov Decision Process (MDP) $(\mathcal{S},\mathcal{A},r,p,\gamma)$ \cite{puterman2014markov}, where $\mathcal{S}$ is the state space\hush{(discrete or continuous)} with $s_t \in \mathcal{S}$, $\mathcal{A}$ is the action space\hush{(discrete or continuous)} with $a_t \in \mathcal{A}$, $r_a^s$ is the reward function\hush{(deterministic or stochastic)}, $p(\cdot|s,a)$ is a conditional probability distribution of the form $p(s_{t+1} | s_t, a_t)$ (environment's dynamics), and $\gamma \in (0, 1]$ is a scalar discount factor.

A Reinforcement Learning agent seeks a policy $\pi^*(a_t | s_t)$ that for each state defines a distribution over possible actions. The policy should maximize {\em cumulative reward} over time, i.e., an expectation under the environment's dynamics:

\begin{equation}\label{rl_objective}
J(\pi) = \E \sum_{t=0}^{\infty}{\gamma^{t}r_{a_t}^{s_t} }
\end{equation}
\begin{equation*}
s_0 = s, a_0 = a, s_t \sim p(\cdot| s_{t-1}, a_{t-1}), a_t \sim \pi(\cdot | s_t) 
\end{equation*}
Based on Equation \ref{rl_objective}, the Reinforcement Learning problem can be viewed as an optimization over the space of policies $\pi$:
\begin{equation}\label{rl_optimization}
\begin{split}
\pi^* = \argmax_{\pi} \: \E  \sum_{t=0}^{\infty}{\gamma^{t}r_{a_t}^{s_t} }
\end{split}
\end{equation}
For policies $\pi_{\theta}$ smoothly parameterized by weights $\theta$, Equation \ref{rl_objective} is amenable to gradient descent in terms of 
$\nabla_{\theta} J(\pi_\theta)$, i.e., policy gradients \cite{sutton2000policy, schulman2015trust}. Functional dependency of $\pi$ on $s_t, a_t$ makes it easier to work with stochastic policies.

Additionally, the Reinforcement Learning objective in Equation \ref{rl_objective} can be optimized by accurately estimating a state-action value function \cite{sutton2018reinforcement}, and then using that value function to derive an optimal policy $\pi^*$. The state-action value function of policy $\pi$ is
\begin{equation}\label{q-function}
Q^\pi(s_t, a_t) = \displaystyle \E_{\substack{s_{t} \sim p(s_{t} | s_{t-1}, a_{t-1})\\a_{t} \sim \pi(a_{t} | s_{t})}} \left [\sum_{t=0}^{\infty}{\gamma^t r_{a_t}^{s_t}} \right],
\end{equation}
and we can express $Q^\pi(s_t,a_t)$ 
as follows:
\begin{equation}\label{dp}
    Q^{\pi}(s_t, a_t) = r_{a_t}^{s_t} + \gamma \E_{\substack{s_{t+1} \sim p(s_{t+1} | s_t, a_t)\\a_{t+1} \sim \pi(a_{t+1} | s_{t+1})}} \left [Q^{\pi}(s_{t+1}, a_{t+1}) \right]
\end{equation}
A policy can be defined for a state-action value function by
\begin{equation}\label{impl_policy}
   \pi(a_t | s_t) = \delta(a_t = \argmax_{a_t \in \mathcal{A}} \: Q(s_t, a_t)),
\end{equation}
 and by substituting Equation \ref{impl_policy} into Equation \ref{dp}, we obtain the Bellman optimality equations \cite{Bellman:1957} that characterize optimal policies in terms of optimal Q-functions $Q^* = Q^{\pi^*}$:
\begin{equation}\label{optimal_q}
    Q^*(s_t, a_t) = r_{a_t}^{s_t}  + \gamma \E_{s_{t+1} \sim p(s_{t+1} | s_t, a_t)}\left[\underset{a_{t+1}}{\max} \: Q^*(s_{t+1, a_{t+1}}) \right]
\end{equation}
Therefore, Reinforcement Learning (Equation \ref{rl_optimization}) seeks an optimal policy $\pi^*$ such that $Q^{\pi^*}(s_t,a_t) \geq Q^{\pi}(s_t,a_t)$, $\forall \pi, s_t \in \mathcal{S}, a_t \in \mathcal{A}$.

\subsection{\textit{Off-Policy} Learning}
\label{Off-policy}

In the discussion below, we focus on policies defined by state-action value functions per Equation \ref{impl_policy}, and $\varepsilon$-greedy policies \cite{sutton2018reinforcement}  that perform random exploration with a small probability. This focus simplifies the narrative, and we found such policies successful in practice. Sections \ref{sec:formal} and \ref{sec:empirical} also discuss parameterized policies that draw actions from state-conditional probability distributions and are trained to maximize cumulative rewards.


Striving for an optimal Q-function $Q^*$ and therefore an optimal policy $\pi^* = \delta(a_t = \argmax \: Q^*(s_t, a_t)), $ Q-learning \cite{watkins1992q} alternates between two phases. First, it improves an approximate estimate $Q(s_t, a_t; \theta)$ of $Q^*$ by repeatedly regressing \hush{the left-hand side and right-hand side of}Equation \ref{optimal_q} with respect to parameters $\theta$. In off-policy Q-learning, an action with the best Q-value is considered Bellman-optimal per Equation \ref{optimal_q} and used in gradient calculations. In the second phase, the agent explores  the environment, typically following a stochastic policy based on $Q(s_t, a_t; \theta)$, e.g., an $\epsilon$-greedy \cite{sutton2018reinforcement} version of $\pi(s_t) = \delta(a_t = \argmax \: Q(s_t, a_t; \theta))$. 
Q-learning takes a single gradient step per training iteration to minimize the difference between the left-hand and right-hand side of Equation \ref{optimal_q}. Among the many variants of this learning procedure, the most common variant called DQN \cite{mnih2013playing} utilizes a replay buffer $\mathcal{D}=\{(s^i_t,a^i_t,s^i_{t+1},r^i_{t+1})\}$ for storing the agent's interaction with the environment and alternates between data collection and gradient steps with respect to the Temporal Difference loss function $\mathcal{L}_i(\theta_i)$,
defined as follows via Equation \ref{optimal_q}:
\begin{equation}\label{q_learning_objective}
\mathcal{L}_i(\theta_i) = \E_{s_t, a_t \sim D} \left[(y_i - Q(s_t, a_t; \theta_i))^2 \right]
\end{equation}
Here $y_i = \E_{s_{t+1} \sim D}\left[ r(s_t,a_t) + \gamma \underset{a}{\max} \: Q(s_{t+1},a; \theta^{-}_i))\right]$, $\theta_i$ are the parameters of the approximate Q-function at the iteration $i$ of the learning procedure, and for better stability in the learning procedure the $\theta_i^{-}$ are frozen parameters that are used for estimating the target values, i.e., $y_i$. The $\theta^{-}$ parameters are periodically updated to the current parameters $\theta$ of the parametric Q-function. Variants of DQN decouple the selection from action evaluation (Double DQN \cite{van2015deep}), or approximate the Q-function through dueling network architectures (Dueling DQN \cite{wang2016dueling}) 
to address the overestimation of Q-values in the DQN setting. 

\hush{
Also, this can be extended for the case where the action space $\mathcal{A}$ of each state is not the same through appropriate selection of the best valid action.}

Q-learning and its aforementioned deep reinforcement learning variants
are characterized as \textit{off-policy} algorithms \cite{sutton2018reinforcement}, since the learning target values $y_i$ can be computed without any consideration of the policy that was used to generate the experiences of the replay buffer i.e., $\mathcal{D}=\{(s^i_t,a^i_t,s^i_{t+1},r^i_{t+1})\}$. 

\section{Offline Reinforcement Learning}
\label{sec:offline}

Offline Reinforcement Learning seeks policy $\pi^*$ to 
maximize cumulative reward (Equation \ref{rl_objective}), but avoids {\em live interactions with the environment during training}. It is used when such interactions  can be harmful~\cite{gauci2018horizon}. Hence, a static dataset of state-action transitions $\mathcal{D_{\pi_\beta}}=\{(s^i_t,a^i_t,s^i_{t+1},r^i_{t+1})\}$ logged via behavioral policy $\pi_\beta$ is used.
Web-based services routinely log each user session, but offline RL based on Equation \ref{q_learning_objective} requires {\em redundant} logs, where each row includes the previous state $s_t$ and next state $s_{t+1}$ \cite{gauci2018horizon}. 
Such rows can be used independently and batch-sampled for training.

Offline RL trains on state-action transitions batch-sampled from \textit{training set}
$\mathcal{D_{\pi_\beta}}$ through \textit{off-policy} learning (Section \ref{Off-policy}). Learning relies on RL mechanisms such as optimal state-action values (e.g.,  Equation \ref{q_learning_objective}), but generalization results from supervised learning should apply. After the learned policy $\pi^*$ is evaluated {\em offline} and tuned,  {\em online} evaluation uses live interaction with the environment.%


\textit{Off-policy} Reinforcement Learning algorithms (DQN, Double DQN, and Dueling DQN) that estimate the state-action value function $Q(s_t, a_t; \theta)$ can be directly used offline. However, the {\em online} regime has the benefit of additional data collection via environment interaction (exploration) that helps the agent refine its $Q(s_t, a_t; \theta)$ estimates for high-reward actions. Although \textit{off-policy} learning has shown promising results in offline RL \cite{kalashnikov2018scalable, pmlr-v119-agarwal20c}, the lack of exploration limits agent's learning \cite{kumar2019stabilizing} due to {\em distributional shift} phenomena.

\noindent \textbf{State distributional shift}
affects offline RL algorithms during the agent's exposure on the real-world environment (deployment phase) \cite{kumar2019stabilizing}. The latter occurs as the agent's learned policy $\pi^*$ may follow a systematically different state visitation frequency compared to the one of the \textit{training set, i.e., $\mathcal{D}_{\pi_\beta}$}, that is induced by the behavior policy $\pi_\beta$. In other words, since the agent's goal is to find the best policy $\pi^*$ offline by utilizing the static dataset $\mathcal{D}_{\pi_\beta}$, its learned policy can diverge from the behavior policy $\pi_\beta$, invoking unreasonable actions in out-of-distribution/unseen states.

\noindent \textbf{Action distributional shift}
affects \textit{off-policy} learning algorithms, which are estimating the Q-function, i.e., $Q(s_t, a_t;\theta)$, during the training process as well \cite{wu2019behavior}. In principle, the accuracy of the regression in Equation \ref{q_learning_objective} depends on the estimate of the Q-function for actions that may be outside of the distribution of actions that the Q-function was ever trained on, i.e., right-hand side of Eq. \ref{optimal_q} . 
The action distribution shift is exacerbated by the differences between the agent's learned policy and the behavioral policy $\pi_\beta$ during training. If the agent's parameterized Q-function produces large, erroneous values for out-of-distribution actions, it will further learn to do so. 

Standard RL methods address the distributional shift via ongoing exploration by revising the state-action value function $Q(s_t, a_t;\theta)$. However, offline RL lacks such a feedback loop, and inaccuracies in the state-action value function accumulate during learning.

\noindent
{\bf Prior research} \cite{schulman2015trust, siegel2020keep} focuses on mitigating distributional shift in offline RL by limiting how much the learned optimal policy $\pi^*$ may deviate from the behavior policy $\pi_\beta$. 
This forces the agent's learned policy to stay close to the behavior policy and reduces state distributional shift. Additionally, the latter reduces action distributional shift during offline training as well, as most of the states and actions fed into the right-hand side of Equation \ref{q_learning_objective} are in-distribution with respect to the \textit{training set}. In that case, the action distributional shift errors should not accumulate. As another example, the work in \cite{jaques2019way} regularizes the agent's learned policy towards the behavioral policy by using the Kullback-Leibler (KL) divergence with a fixed regularization weight, while the Maximum Mean Discrepancy with an adaptively trained regularization weight is used in \cite{kumar2019stabilizing}. Moreover, the work in \cite{wu2019behavior, jaques2019way}
defines target values ($y_i$) to
regularize the agent's learned policy towards the behavioral policy and avoid actions inconsistent with the behavioral policy. 

To guide our design decisions for the application in Section \ref{sec:auth}, in
Section \ref{sec:motivate} we introduce
a simplified, intuitive problem environment.
In Section \ref{subsec:ml_approaches}, we use this environment to illustrate important insights on {\em distributional shift} in Offline RL and the impact of behavioral policy's exploration on the quality of trained models.


\section{Application: User Authentication}
\label{sec:auth}
We illustrate personalization in Web-based services with a self-contained application that is representative of Web-based software and widespread. It optimizes long-term rewards and favors reinforcement learning. We formalize the problem in Section \ref{sec:formal}.

\begin{figure}[h]
\vspace{-2mm}
\centering
\includegraphics[trim=60 125 140 215,clip,width=1\linewidth]{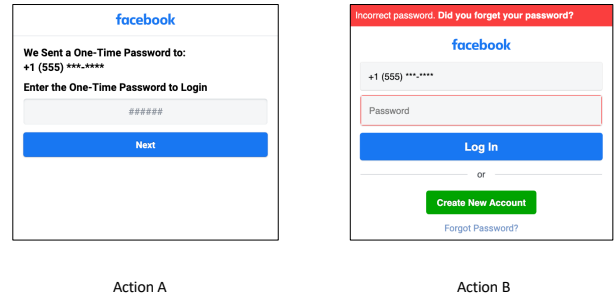}
 \vspace{-6mm}
\caption{\label{fig:auth}User authentication UI.}
\vspace{-2.5mm}
\end{figure}

\noindent
{\bf User authentication} starts a typical Web-based session; thus failures can limit user engagement. Due to forgotten passwords, such failures are common when different passwords are used for each Web-based service. The authentication UI may respond to failures by offering password recovery ({\tt Action A}) or by prompting another log-in attempt 
({\tt Action B}), see Figure \ref{fig:auth}. Password recovery verifies the user's identity by sending a code to a pre-registered mobile phone or another user-owned device. Common options include
\begin{itemize}
    \item One-Time Password (OTP) sent via SMS and email
    \item Time-based One-Time Password (TOTP) authentication \cite{ledesma2020systems}.
\end{itemize}
Such authentication is considered safe unless the user's device is compromised or the phone number is hijacked \cite{sharma2020two}.\footnote{Such authentication can also support
two-factor authentication (2FA) \cite{awasthi2015reducing, rockwell2016two}.} In our application, we authenticate via OTP.

\noindent
{\bf The decision mechanism} that choses between available actions may affect long-term objectives, for example 
{\em user engagement} captured by how many service users are active per day or per month.
To this end, {\tt Action A} may typically succeed and improve user engagement but incurs OTP costs. {\tt Action B} may succeed without such costs, and if it fails again, {\tt Action A} can still be invoked.
\hush{
As a result, the user's daily engagement, which is expressed as a binary value that determines the activeness of the user, i.e., $0$ if the user does not log in for the day, and $1$ otherwise, will be positively impacted. On the other hand, \textit{action a} will imply a corresponding cost (real number) that represents the amount of money that we spent on sending the OTP message to the user.
}
A nascent Web-based service may strive to grow user engagement at greater cost, but mature services may lean towards lower cost. 

\noindent
{\bf Training the decision mechanism on user activity} assumes trends or patterns, as well as sufficient features to learn them. When different user cohorts develop different trends, separating them simplifies the learning task. For example, new users may exhibit different behaviors than long-term users or have less historical data to use for inferences. This paper focuses on users with prior login history and additional user features. 

\noindent
{\bf Additional aspects} of the problem are worth noting. Compared to multiplayer games, the lack of typical-case adversarial behavior simplifies successful policies
and makes them easier to learn from logged data without simulated interactions. On the other hand, training a policy from scratch on user interactions in a live production system may adversely affect user engagement until the policy is optimized. Compared to the (contextual) bandit setup, this problem exhibits sequential depth.

\noindent
{\bf Extensions} of this problem include state-dependent and larger action spaces, and substantial generalizations covered in Section \ref{sec:conclusions}.

\section{Problem formalization}
\label{sec:formal}
The application introduced in Section \ref{sec:auth} deals with sequences of actions, e.g., {\tt Action B} can be tried several times before falling back on {\tt Action A}. The sequential nature of this application
calls for models based on states and state transitions. First, we introduce a simplified, intuitive example used later in the paper and then outline our full-fledged state model.

\subsection{A motivational example}
\label{sec:motivate}
\begin{figure}[h]
\vspace{-2mm}
\centering
\includegraphics[width=0.75\linewidth,keepaspectratio=true]{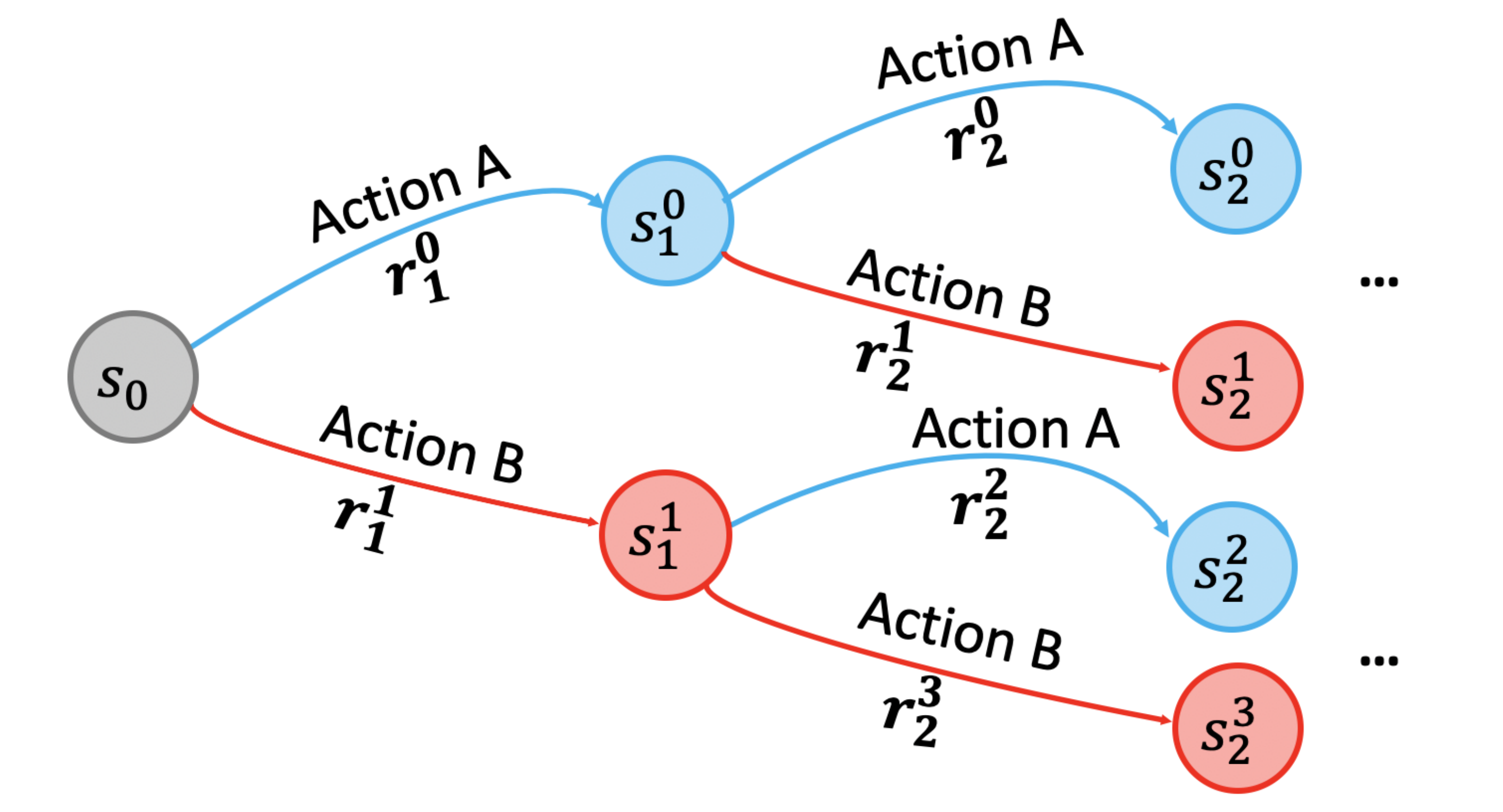}
\vspace{-3mm}
\caption{\label{fig:states} State transitions during user authentication. All states are distinct and indexed with depth (subscripts) and the history of actions (encoded in binary in superscripts).}
\vspace{-2mm}
\end{figure}

Consider the user-authentication application with two possible actions at a time, introduced in Section \ref{sec:auth}. It is easy to check whether {\tt Action A} (password recovery) is more beneficial than {\tt Action B} (another log-in attempt) on average, marginalizing all other information. This generalizes to short sequences, e.g., which action is preferable after one {\tt Action B}? After two of them? (marginalizing everything else). By capturing such sequences up to some depth, one can optimize per-action rewards using either supervised or reinforcement learning. Figure \ref{fig:states} illustrates possible system actions from the initial failed-login state and subsequent states, as well as implied state transitions. The feedback to system actions is captured by rewards only (which include OTP costs as per Equation \ref{eq:reward}), and successful logins bring higher rewards than failures. State transitions depend entirely on the actions chosen by the system. In particular, after a successful login, the sequence can continue upon a future login failure. This self-contained state-reward model can capture the most basic sequential preferences but lacks user personalization, which is key to our work.

A user can be characterized by a number of features that cover, e.g., time since account creation and local time, the amount of online activity, etc.
In other words, each user at each step is represented by a multidimensional vector such that proximal vectors represent similar users. Compared to the earlier few-state model that represents an average user, personalization requires states that cannot be limited to a small predefined set. We now interpret Figure \ref{fig:states} as showing possible transitions and traces for one user, wherein each user state $s_t$ in each trace is represented by a finite-dimensional feature vector (user state vector).

 The semantics of features are not important for our motivational example. Therefore, for each user we represent
 the initial state $s_0$ (the first failed log-in attempt)
 with a state vector by drawing
 it at random from a personalized multivariate Gaussian distribution with mean and covariance selected randomly per user. Subsequent user states $s_t$ with $t = 1, \dots$ that correspond to future failed login attempts are generated randomly and as part of state transitions. Specifically, a base (user) feature vector is drawn from a multivariate Gaussian that is defined per user, per discrete time step $t$, and per action $a_t \in \{A, B\}$ that was prompted by the authentication UI on user state $s_{t-1}$. This base vector and the vector representing user state $s_{t-1}$ are then aggregated to construct the vector for the next user state $s_t$.
 The user engagement and corresponding action costs on each user transition $(s_t, a_t, s_{t+1})$ are represented by a single normalized scalar value-reward $r_{t+1}$. The latter is drawn from a normal distribution with randomly selected mean and variance, defined per user, per time step, per action.

The simulated environment we introduced for system actions during user authentication is used as a testbed for evaluating ML techniques in Section \ref{sec:apply}.
Under our formulation, the trained decision mechanism that chooses an action for each transition aims to optimize and/or balance long-term objectives, i.e., cumulative rewards over a time horizon (Section \ref{subsec:rl}).  

\subsection{Our practical state model and rewards}
\label{subsec:state}
The problem formalization we use in practice
 differs from the motivational example in two aspects: (1) user state vectors are based on actual user features rather than drawn at random, (2) rewards are crafted to approximate long-term objectives.
 
\noindent
{\bf User features} 
for state representation include {\em real-time (contextual)} and {\em historical} features. On-device real-time features capture the app platform, local time of day, day of the week,
login time, etc. Historical features focus on past events, such as the total number of user password recoveries sent in the past, time since last login, etc. Historical features may also aggregate user data specific to the authentication process, e.g.,
the number of authentication attempts in the last 30 days and their outcomes, 
which of {\tt Action A} and {\tt Action B} were used, and how the user reacted.

\noindent
{\bf Rewards} 
are formulated to help optimize long-term objectives that are affected by the authentication decision mechanism. Specifically, we aim to optimize {\em user engagement} and {\em monetary cost}. In our Web-based service, user engagement is reported and expressed as a binary value that determines if the user was active during the day. Monetary cost represents the amount of money spent on password recovery ({\tt Action A})  after failed login attempts. Ideally, we would like to avoid harming user engagement and not block users from logging in, while reducing charges for {\tt Action A}. To this end, we formulate the reward as a linear combination of user engagement (UE) and monetary cost (Cost), thus
\begin{equation}
\vspace{-1mm}
\label{eq:reward}
r = w_{u}\cdot \mathrm{UE} - w_{c} \cdot \mathrm{Cost}, 
~\mathrm{where}~w_u, w_c \in \mathbb{R}
\end{equation}

\begin{figure}[b]
\centering
\vspace{-3mm}
\includegraphics[width=0.95\linewidth,keepaspectratio=true]{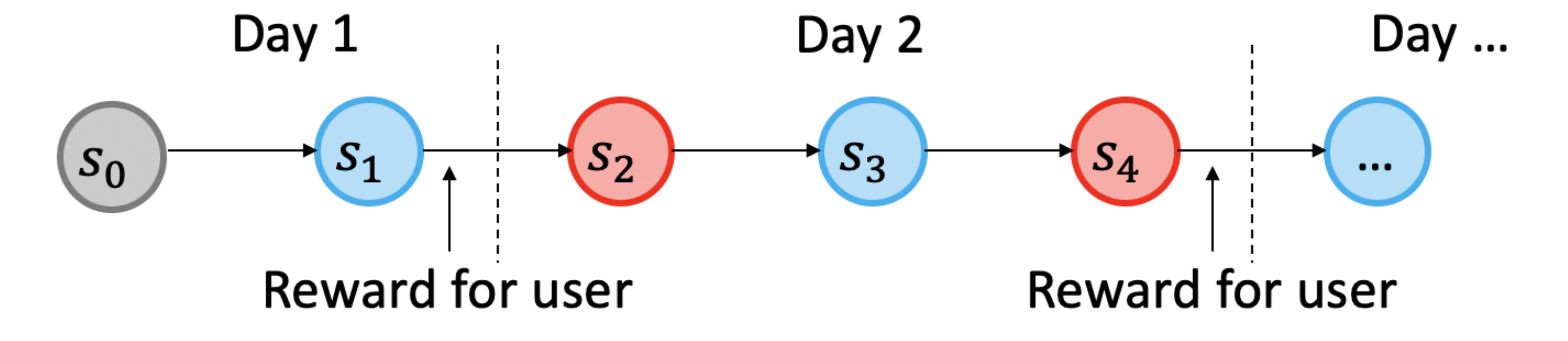}
\vspace{-3mm}
\caption{\label{fig:rewards}
 Rewards are set daily per user, to the last action.
}
\end{figure}

Varying the weights helps explore the Pareto curve of multiobjective tradeoffs relevant to the application. \hush{We noticed that assigning rewards to each action increases variance and encourages unnecessary password recovery attempts.} We assign rewards to the last action on each day (Figure \ref{fig:rewards}) because system metrics for user engagement are tallied at the end of each day. During RL training, rewards are propagated back via the Bellman equation. 

\begin{figure} [t]
\centering
\includegraphics[width=0.81\linewidth,keepaspectratio=true]{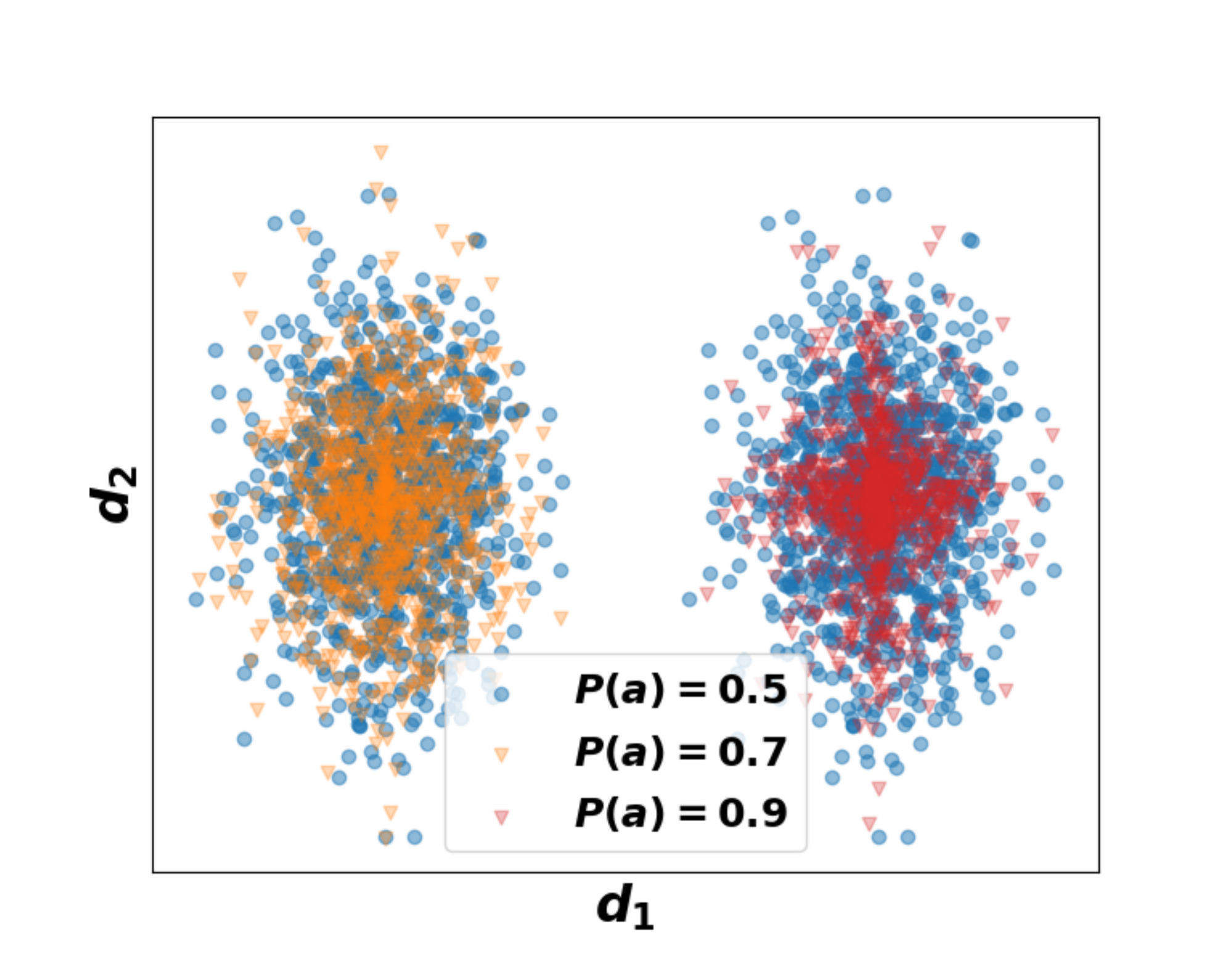}
\caption{\label{fig:coverage}Exploration diversity improves state coverage. To visualize distributions of visited states in the feature space, we use two principal components via PCA. P$(A)=0.5$ exhibits the greatest state coverage (blue), whereas P$(A)=0.9$ exhibits 
limited coverage (red). The blue pointset is cloned.
}
\end{figure}

\hush{
\noindent
Large-scale Web-based systems optimize multiple metrics/objectives, and this can be viewed as a two-stage process --- first, improve individual objectives without negative impact on other objectives, and then explore tradeoffs between multiple objectives based on the business context and needs.
The use of weighted sums primarily targets the second stage, where varying weights can help tabulate the Pareto curve for multiple objectives.
In practice, individual terms in Equation
\ref{eq:reward} and other such expressions should be comparable in magnitude so that no one term dominates the linear combination and the objective remains sensitive to each term. Results reported in Section \ref{sec:empirical}
include weight tuning.
}

\section{Applying Reinforcement Learning}\label{sec:apply}
 The problem formalization in Section \ref{sec:formal} facilitates established ML techniques, including {\em supervised} and {\em reinforcement learning}. In this section we discuss the pros and cons of competing ML approaches, introduce our proposed RL model, then explain how training data is extracted and augmented.
 
 \subsection{Competing ML approaches}\label{subsec:ml_approaches}

\begin{figure}[t]
\centering
\includegraphics[width=0.82\linewidth,keepaspectratio=true]{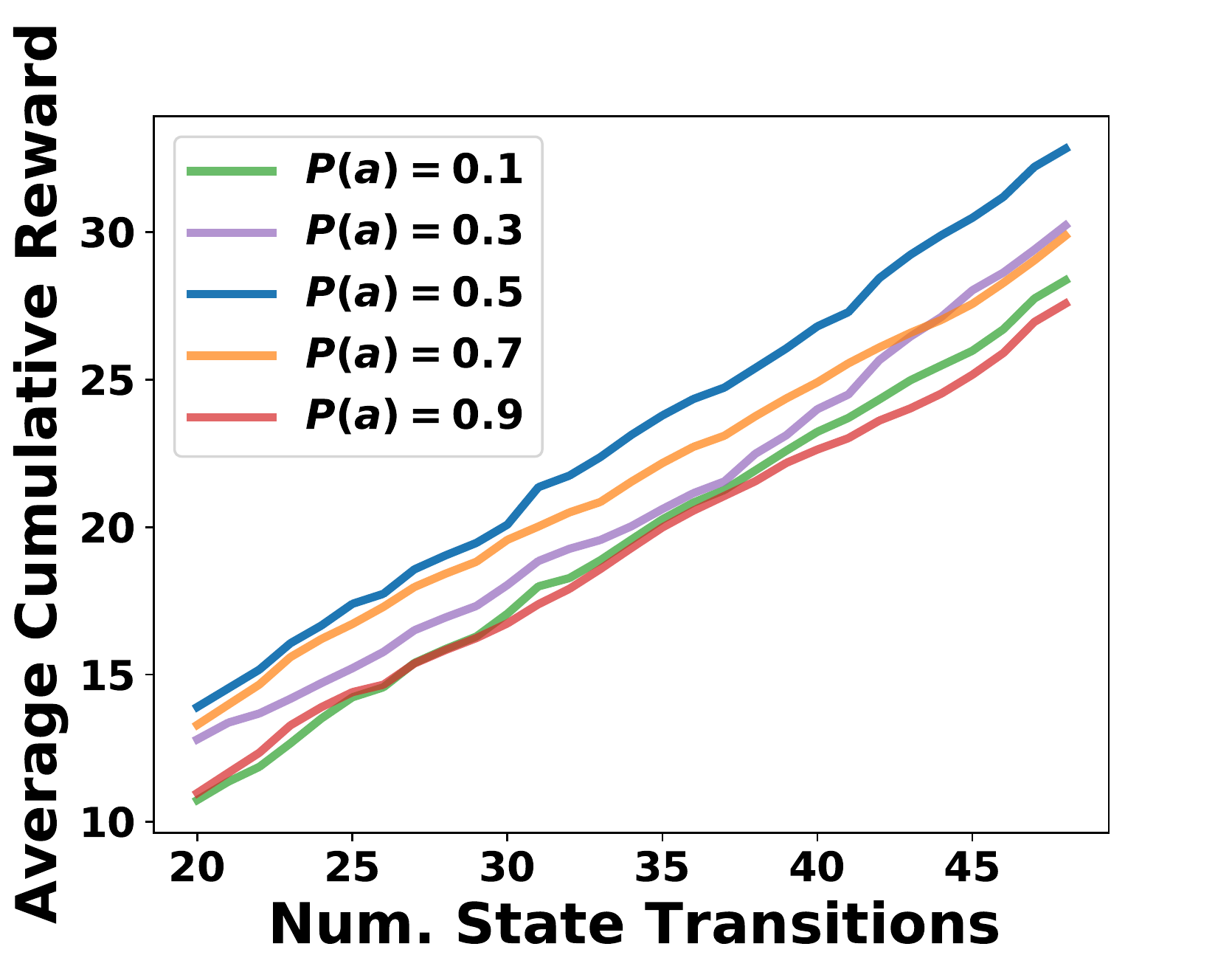}
\caption{\label{fig:trajectories}The impact of exploration diversity on learning returns. Between {\tt Actions A} and {\tt B}, the former is chosen with probability P$(A)$. The plot shows average cumulative rewards over trajectories of different length. The best results are observed for balanced exploration P$(A)=0.5$.}
\vspace{-2mm }
\end{figure}

 For a given state vector in the feature space, an ML model can ($i$) predict rewards for each action via Supervised Learning or ($ii$) estimate Q-values via offline Reinforcement Learning (Section \ref{sec:offline}). Generally, Reinforcement Learning offers two important advantages:
 \begin{itemize}
     \item RL exploration policies are linked to exploitation and optimized to improve state coverage.
     \item Long-term cumulative objectives, e.g., user engagement, can be optimized by RL even when they are not faithfully represented by immediate rewards for each action (Figure \ref{fig:rewards}).
 \end{itemize}
 The impact of the exploration level of behavioral policies in offline RL is illustrated in Figure \ref{fig:coverage} that is produced for the motivational example from Section \ref{sec:motivate}. Here we compare (behavioral) exploration policies defined by the complementary probabilities of {\tt Actions A} and {\tt B}. Specifically, using the induced state transitions and corresponding rewards as a static dataset, we train offline an RL model for each behavioral policy  (Section \ref{sec:offline}) through an established \textit{off-policy} algorithm (DQN). Intuitively, balanced exploration with equal probabilities should provide better state coverage. Indeed, this is observed in our simulation. In the figure, we project state vectors from the feature space onto two principal components (via PCA). The yellow (70:30) and red (90:10) point clouds are narrower than the blue (50:50) point cloud. Furthermore, Figure \ref{fig:trajectories} shows that exploration diversity in the behavioral policy helps train a better model. In other words, during the evaluation phase, we can expect
 higher perceived returns (cumulative rewards) for the RL model trained offline with balanced exploration. Such evaluation is performed on users not seen in the training dataset and by using the trained RL model as a decision mechanism that drives the state transitions. The impact of \hush{exploration due to}behavioral policies' exploration is related to the {\em distributional shift} phenomemon discussed in Section \ref{sec:offline}. The higher the state coverage in the training dataset, the less significant the state distributional shift during evaluation and action distributional shift during training for an Offline RL model.
 
 Modeling {\em forward sequential depth} (i.e., the future impact of current actions) is difficult for supervised learning. Modeling {\em backward sequential depth} (the impact of past actions) is somewhat easier, using additional features that summarize past states and actions. Furthermore, reinforcement learning often suffers from high variance in rewards and statistical bias in explored states.
 
 Despite obvious limitations, supervised learning (SL) remains a viable competitor to RL because
 \begin{itemize} 
 \item 
 SL often facilitates more mature optimization methods
 \item 
 SL can train higher-quality models faster due to smaller variances and greater learning rates. 
 \end{itemize}
 Therefore, to empirically evaluate our proposed RL method, Section \ref{sec:empirical} compares it to a production system trained with supervised learning. Among different RL techniques,
 {\em model-based} RL (tree search, etc), runs into difficulties 
 predicting state transitions based on actions and corresponding rewards (environment dynamics).

\subsection{RL decision models}
\label{subsec:decision_models}
Given the large state space in our application, we model the Q-function by a neural net.
The values of the trained $Q(s_t, a_t; \theta)$ express accumulated rewards (Equation \ref{eq:reward}) for {\tt Actions A, B} based on user features for a given state. We perform training offline on a static dataset of state transitions using the DQN algorithm. As a result, the task of learning a near-optimal Q-function is transformed into a familiar regression problem for Equation \ref{q_learning_objective}. A learned near-optimal Q-function supports a decision mechanism that selects {\tt Action A} or {\tt B} based on current user features to maximize accumulated rewards. The use of {\em Temporal Difference} (Equation \ref{q_learning_objective}) as the loss function is a key distinction from supervised learning. The implied use of the Bellman equation is what allows this approach to track long-term rewards.

Our implementation (Section \ref{sec:empirical}) makes it easy to try more sophisticated RL models and algorithms with
the same state model and rewards (Section \ref{subsec:state}).
In fact, we have tried several improvements to DQN (such as DDQN, Dueling DQN) and concluded that DQN does not leave much
room for improvement despite being simpler. However, more techniques that parameterize both the policy and the Q-function are generally worth trying.
Whether or not this improves long-term objectives is unclear {\em a priori}, and we therefore implement a recent technique in this category.

 \textit{Critic-Regularized Regression} (CRR) \cite{wang2020critic} is an off-policy method that discourages sampling low-quality actions from the training dataset. As it is typical for actor-critic algorithms, CRR parameterizes both the policy and the Q-function \cite{NIPS1999_6449f44a}. It additionally transforms Q-values in the objective (Equation \ref{rl_objective}) with a monotonically increasing function, such as $\exp(\cdot)$, to emphasize higher Q-values. This way, during policy update, $\pi$ more often samples high-quality actions within the training distribution. The critic is trained using distributional Q-learning \cite{wang2020critic, barth2018distributed}. The distributional representation of the returns translates well into stochastic behavioral policies.

\subsection{Preparing data for training}\label{subsec:data_preperation}
Each row of our training data includes user features, actual system actions,
and rewards.

\noindent
{\bf Data extraction.}
Motivated by the impact of the exploration level of behavioral policies in Offline RL (Section \ref{subsec:ml_approaches}) we choose equal probabilities for {\tt Actions A} and {\tt B}. Thus, upon a failed login attempt, an action is chosen at random. Another handler starts extracting and computing user features simultaneously. The two asynchronous events are tagged with the same unique ID and joined by a stream processing system for logging in the training table. The user engagement metric and the sum of OTP fees are computed by the end of the day. These rewards are joined with the training table based on an anonymized user ID column, where rewards are attributed to the last login failure event of the day (see Figure \ref{fig:rewards}). We logged data using the described exploratory behavioral policy for a time period of three weeks.

\noindent 
{\bf Data augmentation.}
Following the standard Markov Decision Process (MDP) framework, RL models are trained on consecutive pairs of state/action tuples that correspond to state transitions in user sequences (Figure \ref{fig:states}). We use the open-source applied Reinforcement Learning platform \textit{ReAgent} (Horizon) \cite{gauci2018horizon} to transform logged state-action-reward data to the following row format:
\begin{itemize}
    \item {\tt MDP ID}: a unique id, e.g., anonymized user ID, for the Markov Decision Process chain.
    \item {\tt Sequence Number}: a number representing the depth of the state in the MDP, e.g., the timestamp of the state.
    \item {\tt State Features}: the user features for state representation of the current step.
    \item {\tt Action}: the actual system action.
    \item {\tt Action Probability}: the probability that the current system took the action logged.
    \item {\tt Metrics}: user engagement metric and the sum of OTP fees.
    \item {\tt Possible Actions}: an array of possible actions at the current step.
    \item {\tt Next State Features}: the user features for state representation of the subsequent step.
    \item {\tt Next Action}: the actual system action at the next step.
    \item {\tt Sequence Number Ordinal}: a number representing the depth of the state in the MDP after converting the Sequence Number to an ordinal number.
    \item {\tt Time Diff}: a number representing the time difference between the current and next state.
    \item {\tt Possible Next Actions}: an array of actions that were possible at the next step.
\end{itemize}
The metrics map enables reward shaping for Equation \ref{eq:reward} by tuning $w_u, w_c$ during training. Offline RL training is enabled by preprocessing logic for the row format above.

\section{Empirical evaluation}
\label{sec:empirical}

To evaluate ML techniques from Section \ref{sec:apply},
\hush{to the application from Section \ref{sec:auth} that we formalized in Section \ref{sec:formal}}
our workflow first assesses quality of RL models trained offline. Hyperparameter tuning is performed based on offline metrics. For the best seen RL model, online evaluation on live data helps us compare to a baseline production solution that uses SL. Our evaluation methodology and especially the metrics are applicable beyond this work.

\subsection{Offline training and evaluation for RL}

To support decision models from Section \ref{subsec:decision_models}, we train our DQN model using the ReAgent (Horizon) platform \cite{gauci2018horizon} and its default neural network architecture. We set the discount factor to $\gamma=1.0$ because most MDP sequences are short (80\% of them have a single step). Equation \ref{eq:reward} is used for the reward function.

\begin{figure}[b!]
\centering
\vspace{-5mm}
\includegraphics[width=1\linewidth,keepaspectratio=true]{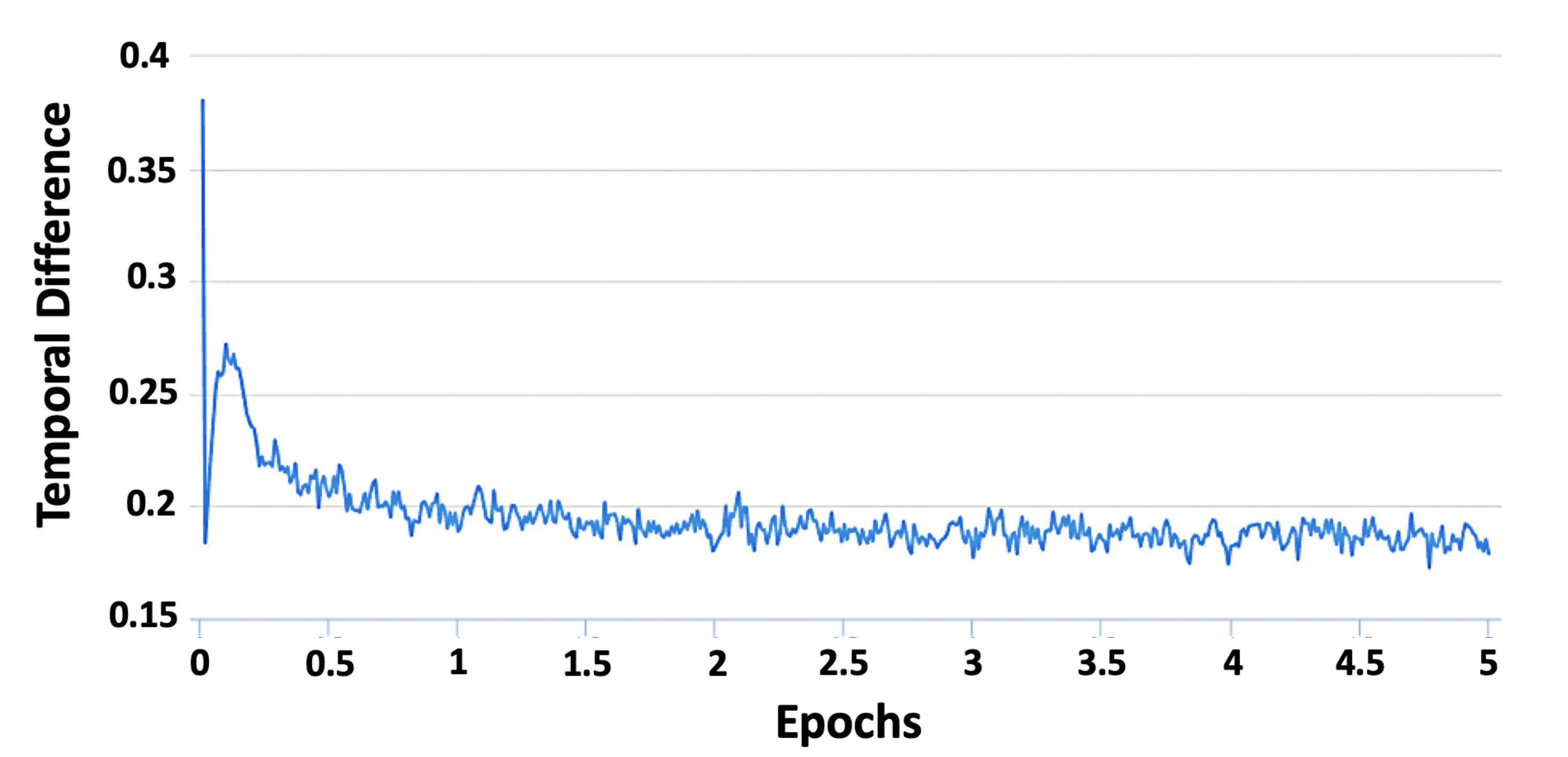}
\vspace{-5mm}
\caption{\label{fig:td_loss}Temporal Difference loss (MSE) with respect to learning epochs for the DQN RL model (Section \ref{subsec:decision_models}).
}
\vspace{-1mm}
\end{figure}

\noindent
{\bf Offline RL training} minimizes {\em Temporal Difference} loss (Equation \ref{q_learning_objective}) over the provided static dataset of transitions (Section \ref{subsec:data_preperation}).  Figure \ref{fig:td_loss} shows how loss values
for the DQN RL model (Section \ref{subsec:decision_models}) change
over training epochs that consist of mini-batches (iterations). 
To avoid overfitting, we split the static dataset of transitions into a training and test set, then compare the average Q-values of the same {\tt Actions A, B} between the training and test sets, and keep the differences below $10\%$. 

\noindent
{\bf Offline evaluation} avoids the dangers of poorly trained policies in production. Additionally, it uses a static dataset for consistent experimentation and performance tuning. To watch out for deficient training and distributional shifts (Section \ref{sec:offline}), our evaluation methodology pays particular attention to training quality metrics. 

\begin{figure}[b!]
\centering
\vspace{-3mm}
\includegraphics[width=1\linewidth,keepaspectratio=true]{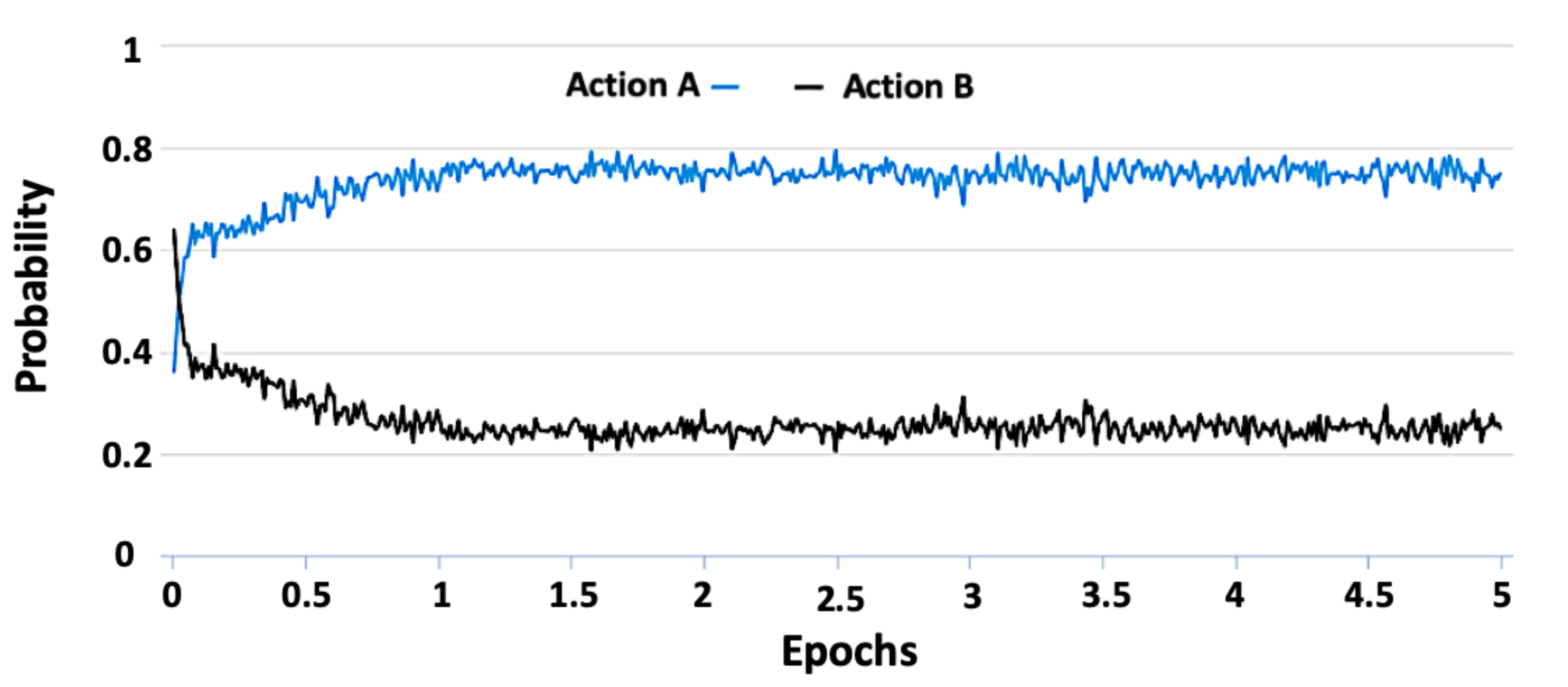}
\vspace{-7mm}
\caption{\label{fig:action}Changes in action distributions for {\tt Actions A, B} over learning epochs for the DQN RL model (Section \ref{subsec:decision_models}).
}
\end{figure}

\begin{itemize}
    \item {\em The Temporal Difference Ratio}
    addresses a pitfall in using raw {\em Temporal Difference} (TD) loss (Equation \ref{q_learning_objective}), which is sensitive to the magnitude of Q-values.
    A poor RL model will exhibit low TD values
    when Q-values are low. Instead, we evaluate the metric $\frac{\mathrm{TD}}{\min \{\bar{Q}_A, \bar{Q}_B\}}$, where $\bar{Q}_A, \bar{Q}_B$ are the average Q-values of {\tt Actions A, B} over the training dataset.
    \item {\em Distributional stability} during training is vital to effective learning. The action distribution of the trained offline RL model should not shift too far from the behavioral policy, which we measure
    as the KL-divergence \cite{jaques2019way} 
    between the RL model's learned policy and the behavioral policy function. 
    The action distribution of our DQN model in Figure \ref{fig:action} stabilizes over training. 
    The change of the action distribution over a time window ($25$ training iterations) is a good proxy for the distributional stability for offline RL models.
\end{itemize}

\noindent
{\bf Counterfactual Policy Evaluation (CPE)} with respect to {\em perceived accumulated reward} is the second tier of our evaluation methodology. CPE is performed
{\em offline} (because our RL model is trained offline) and {\em off-policy} because it uses a static dataset of transitions $\mathcal{D_{\pi_\beta}}$, logged by the behavioral policy $\pi_\beta$. The key idea is to answer counterfactual questions of the form: ``how probable would it be for the offline policy to encounter/perceive same state transitions/rewards as the logged ones?". To this end, CPE provides a safe way of ranking different offline RL models before they are tested in production. Off-policy evaluation
can be based on ordinary importance sampling \cite{importance_sampling}, 
but at risk of high variance which grows exponentially with the number of state transitions. More sophisticated methods \cite{off_policy_eval} provide controlled bias-variance tradeoffs by utilizing additional information, e.g., the learned Q-function. In our application, relatively short MDP sequences make it practical to use the {\em sequential doubly robust estimator} \cite{doubly_robust} to produce unbiased performance estimates for policies trained offline.


\noindent
{\bf Hyperparameter optimization} for Offline RL is challenging \cite{paine2020hyperparameter}, as there is no access to the environment.
Hyperparameters can be optimized toward the metrics in our two-tier offline evaluation above, after which the best seen models can be evaluated online. For example, parameters for training, such as the learning rate, can be optimized via the first tier, whereas the second tier can compare different RL algorithms. Our DQN model (Figures \ref{fig:td_loss} and \ref{fig:action}) exhibited the best results in offline evaluation, was deployed online and evaluated vs. SL baseline production system (Section \ref{subsec:empirical}).

\subsection{Empirical comparisons}
\label{subsec:empirical}
\noindent
{\bf Our baseline using supervised learning} is
a prior production model with a tree-based meta-learner that estimates the Conditional Average Treatment Effects \cite{kunzel2019metalearners}. The treatment effects are represented by a linear combination of the rewards defined in Equation \ref{eq:reward}. The meta-learner decomposes the decision process into two steps by first estimating the conditional reward expectations, $U = \E[\mathrm{UE} \mid a_t, s = s_t]$, $C = \E[\mathrm{Cost} \mid a_t, s = s_t]$, where $a_t \in A$.  $U$ and $C$ are computed using Gradient Boosted Decision Trees whose parameters, including learning rate, tree depth and tree leaves, are swept in a large range to minimize mean squared errors. Then the learner takes the differences between the estimates and chooses the action that gives the highest reward:
\begin{equation}
a_t =\argmax_{a_t \in A}
\E[r_t | a_t, s = s_t] 
\end{equation}
After deriving $r_t$ using Equation \ref{eq:reward}, the final action is given by
\begin{equation}
a_t = \argmax_{a_t \in A} (w_u \cdot U - w_c \cdot C)
\end{equation}
The reward weights $w_u$ and $w_c$ are chosen through online experiments to ensure compelling
multiobjective tradeoffs.

\begin{table}[tb]
  \caption{\label{tab:online} 
  SL and RL policies are compared online to fixed policies that always perform {\tt Action A} or {\tt Action B}. RL provides the most attractive tradeoffs, shown in bold.
  \vspace{-2mm}}
  \begin{tabular}{lccc}
    \toprule
     & \multicolumn{2}{c}{\sc User engagement} & \sc OTP cost  \\
     Policy\hspace{-4mm} & \sc Daily & \sc Monthly &   \\ 
    \midrule
    {\tt  B} & - & - & -  \\
     {\tt  A}  & +1.49\% $\pm$ 0.273\%   &  +2.81\% $\pm$ 0.279\% & +120\% $\pm$ 4.16\%  \\
    SL  & +1.35\% $\pm$ 0.212\%  & +2.58\% $\pm$ 0.116\% & +81.1\% $\pm$ 3.13\%   \\
    RL & \bf+1.50\% $\pm$ 0.213\%    & \bf +2.55\% $\pm$ 0.183\% & \bf +70.3\% $\pm$ 2.96\% \\
  \bottomrule
\end{tabular}
\vspace{-2mm}
\end{table}

\begin{table}[b]
  \vspace{-2mm}
  \caption{\label{tab:action_dist} Action distributions (\% of {\tt Action A}) for behavioral policies trained with SL and RL by user cohort.\vspace{-2mm}}
  \begin{tabular}{lcc}
    \toprule
     & SL & RL   \\
    \midrule
    All users &  55.88\%  &  22.75\% \\
    Single-login users &   60.16\% & 36.41\%\\
    Multiple-login users & 49.40\% & 11.19\%\\
  
  \bottomrule
\end{tabular}
\end{table}

\noindent
{\bf Online evaluation} is performed using one-month's data. Table \ref{tab:online} demonstrates the first round of results at a 95\% confidence level, where our RL model is trained on randomized data as per Section \ref{subsec:data_preperation}.
The RL model is compared to fixed policies (the same action repeated always) and a prior production model based on SL.
The RL model significantly reduces OTP costs while exhibiting competitive daily and monthly user engagement results. We estimate \textit{Return On Investment} (ROI) by dividing total cost by total monthly engagement. RL outperforms SL by $5.93\% \pm 0.782\%$. To collect data for {\em recurrent training}, we deploy our RL agent by extending the deterministic policy to an $\epsilon$-greedy policy with $\epsilon = 0.1$.

When ML is applied to practical problems, it commonly
optimizes {\em surrogate objectives}, and one has to empirically
check that practical objectives are improved as a result. Moreover, sophisticated applications often track multiple performance metrics whose regressions may block new 
optimization. In our case, one such metric counts {\em notification disavow events} (NDEs) for password reset, where a user turns off notifications, perhaps because there were too many. 
Comparing to SL, we observe that RL reduces NDEs by 50\%.
To this end, Table \ref{tab:action_dist} shows that the RL agent trades off {\tt Action A} for {\tt Action B} and thus reduces authentication messages, while maintaining neutral engagement metrics. We also group users into cohorts ---
{\em single-login users} and {\em multiple-login users} (within the evaluation period). Intuitively, multiple-login users are more likely to enter their password correctly without help from our authentication messages. From Table~\ref{tab:action_dist}, we see that both RL and SL make use of this user attribute but RL exploits it more efficiently

\begin{figure}[tb]
\centering
\includegraphics[width=0.8\linewidth,keepaspectratio=true]{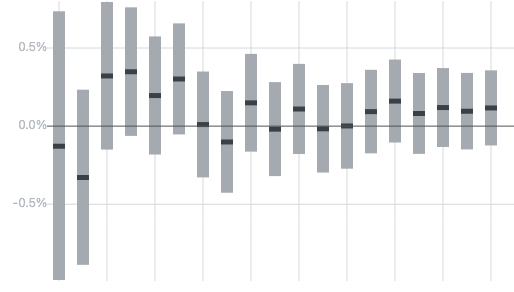}
\vspace{-3mm}
\caption{\label{fig:daily_effect} Daily user engagement of a recurrent CRR model
}
\vspace{-1mm}
\end{figure}

\noindent
{\bf Recurrent training and stability evaluation.} When user behaviors shift over time, ML models must be refreshed using the most recent data (recurrent training). The initial RL model trained on randomized data shows sizable improvements. However,  continued collection of randomized data is risky in terms of direct costs and user experience. The "Rand" policy (Section \ref{subsec:ml_approaches}) in Table \ref{tab:recurrent_result} suggests that its deployment limits the benefits of RL models. Thus, we refresh the model using behavioral data collected from our RL agent. 
Additionally, to check if recurrent training stabilizes in the long term compared with the initial RL model, we perform a second round of online evaluation to study recurrent training using RL behavioral  data and model stability in the long term.

We train a DQN and a CRR model (Section \ref{subsec:decision_models}) with the same parameters and rewards as the initial RL model, but collect their training data via policies from the deployed RL agent. Results summarized in Table \ref{tab:recurrent_result} use the same settings as before (a 95\% confidence level, one-month's testing window, and an $\epsilon$-greedy policy on DQN and CRR with 10\% exploration rate). Both recurrent DQN and CRR give neutral results compared with the initial RL model, which indicates stable performance in recurrent training. Figure \ref{fig:daily_effect} additionally shows daily user engagement for the CRR recurrent model vs. the initial RL model, and this also shows consistent performance.

\begin{table}[b]
  \caption{\label{tab:recurrent_result} Recurrent training results compared with the initial RL agent trained on randomized data (the "Rand" policy issues {\tt Action A} and {\tt Action B} with equal probability).
  }
  \begin{tabular}{lccc}
    \toprule
     & \multicolumn{2}{c}{\sc User engagement} & \sc OTP cost \\
     Policy\hspace{-4mm} & \sc Daily & \sc Monthly &   \\
    \midrule
     DQN  & +0.0023\% $\pm$0.106\%  &  +0.0576\% $\pm$0.0841\% &  -0.166\% $\pm$0.439\%  \\
    CRR  & +0.116\% $\pm$0.240\%  & +0.107\% $\pm$0.140\%   & +1.27\% $\pm$2.87\%  \\
    Rand & 
   -0.336\% $\pm$0.266\%  &  -0.693\% $\pm$0.162\% &  +3.78\% $\pm$3.29\% \\
  \bottomrule
\end{tabular}
\end{table}

\section{Conclusions}
\label{sec:conclusions}

 In this work we show how to apply reinforcement learning (RL) to personalize user authentication in a Web-based system and compare RL to a competing approach based on supervised learning. Working with industry data, using {\em offline} RL avoids releasing poorly trained agents in production. However, this approach requires careful extraction and augmentation of training data to ensure that
 off-policy learning does not succumb to {\em distributional shift}. In practice, RL is often susceptible to high variance and high bias at several of its stages, especially when operating on large-scale live data. Fortunately, our method is sufficiently robust
 for production deployment using the {\em ReAgent} (Horizon) platform \cite{gauci2018horizon}.
 Starting from state modeling and data collection, we articulate obstacles and milestones toward model training and demonstrate practical improvement 
 in end-to-end system-level metrics at a large-scale deployment at Facebook. During development, we use a simplified problem environment to test intuition without sensitive data. 

 \noindent
 {\bf Broader applications.}
 Our self-contained didactic application is not only critical to many Web-based systems, but also generalizes well (e.g., to more than two actions at a time) and illustrates how other aspects of Web-based systems can be personalized and enhanced via ML. Relevant optimizations explored previously include
 \begin{itemize}
     \item personalized individual user notifications \cite{gauci2018horizon},
     \item page prefetching to optimize user experience \cite{MLSYS2020_8f53295a}.
 \end{itemize}
 User notifications and product delivery must balance utility with distraction, while page prefetches improve access latency at the cost of increased network bandwidth.
 These applications bear structural similarity to our work and share several salient aspects.
\begin{enumerate}
\item At each step, the system chooses from several actions. Future decisions are improved based on user feedback.
\item One must optimize and/or balance long-term cumulative objectives, some of which do not reduce to rewards for individual actions that can be handled by supervised learning.
\item Personalization is based on a number of user features and can be supported by ML models in the feature space.
\end{enumerate}

 \noindent
 {\bf More general RL methods.}
 Our straightforward RL model (DQN) is facilitated by the small action space, which ($i$) is less demanding in terms of learning robustness, and ($ii$) allows for simpler neural-net representations where separate outputs produce $Q(s,a)$ values for different actions $a$.
  However, the overall approach generalizes to more sophisticated settings and can be extrapolated to other personalized enhancements for Web-based systems as follows.
  \begin{itemize}
  \item We use model-free RL without predicting future states and action rewards, but such predictions can be leveraged.
  \item Our extraction and handling of user features, as well as offline model-free training and the evaluation methodology are not tied to specific RL models.
  \item Our use of the open-source {\em ReAgent} (Horizon) platform \cite{gauci2018horizon} makes it easy to employ models such as Double DQN \cite{van2015deep} (DDQN) and Dueling DQN \cite{wang2016dueling} that could provide a more stable learning for larger action spaces.
  \end{itemize}
  For very large or continuous action spaces, $Q(s,a)$ can be modeled as a function of both $s$ and $a$ \cite{lillicrap2019continuous}. Continuous action spaces can be addressed using {\em policy gradient} methods \cite{sutton2000policy}, available in {\em ReAgent}.



\newpage
\bibliographystyle{ACM-Reference-Format}
\bibliography{main}



\end{document}